\documentclass[final]{cvpr}

\usepackage{times}
\usepackage{epsfig}
\usepackage{graphicx}
\usepackage{amsmath}
\usepackage{enumerate}
\usepackage{makecell}

\usepackage[noend]{algpseudocode}
\usepackage[]{algorithm2e}
\usepackage{caption}
\usepackage{subcaption}
\usepackage{float}
\usepackage{dblfloatfix}
\usepackage[font={small}]{caption}
\usepackage{amssymb}
\usepackage{pifont}

\usepackage{multirow}
\usepackage{booktabs}
\usepackage[dvipsnames,table]{xcolor}

\usepackage[pagebackref=true,breaklinks=true,colorlinks,bookmarks=false]{hyperref}

\def\cvprPaperID{} 
\def\confYear{CVPR 2021}

\begin{document}

\title{Simple Training Strategies and Model Scaling for Object Detection}


\author{
Xianzhi Du\thanks{Authors contributed equally.}\hspace{2mm}$^1$\qquad
Barret Zoph\footnotemark[1]\hspace{2mm}$^1$\qquad
Wei-Chih Hung$^2$\qquad
Tsung-Yi Lin$^1$\\
$^1$Google\qquad
$^2$Waymo
\\
{\tt\small \{xianzhi,barretzoph,hungwayne,tsungyi\}@google.com}
}

\maketitle
\thispagestyle{empty}

\begin{abstract}

The speed-accuracy Pareto curve of object detection systems have advanced through a combination of better model architectures, training and inference methods.
In this paper, we methodically evaluate a variety of these techniques to understand where most of the improvements in modern detection systems come from. We benchmark these improvements on the vanilla ResNet-FPN backbone with RetinaNet and RCNN detectors. The vanilla detectors are improved by 7.7\% in accuracy while being 30\% faster in speed. 
We further provide simple scaling strategies to generate family of models that form two Pareto curves, named \textbf{RetinaNet-RS} and \textbf{Cascade RCNN-RS}. These simple rescaled detectors explore the speed-accuracy trade-off between the one-stage RetinaNet detectors and two-stage RCNN detectors.
Our largest Cascade RCNN-RS models achieve 52.9\% AP with a ResNet152-FPN backbone and 53.6\% with a SpineNet143L backbone.
Finally, we show the ResNet architecture, with three minor architectural changes, outperforms EfficientNet as the backbone for object detection and instance segmentation systems. Code and checkpoints will be released \footnote{Code and checkpoints will be available in TensorFlow: \\ \url{https://github.com/tensorflow/models/tree/master/official/vision/beta}\\ \url{https://github.com/tensorflow/tpu/tree/master/models/official/detection}}.
\end{abstract}

\begin{figure}[h]
  \centering
  \includegraphics[width=1.0\columnwidth]{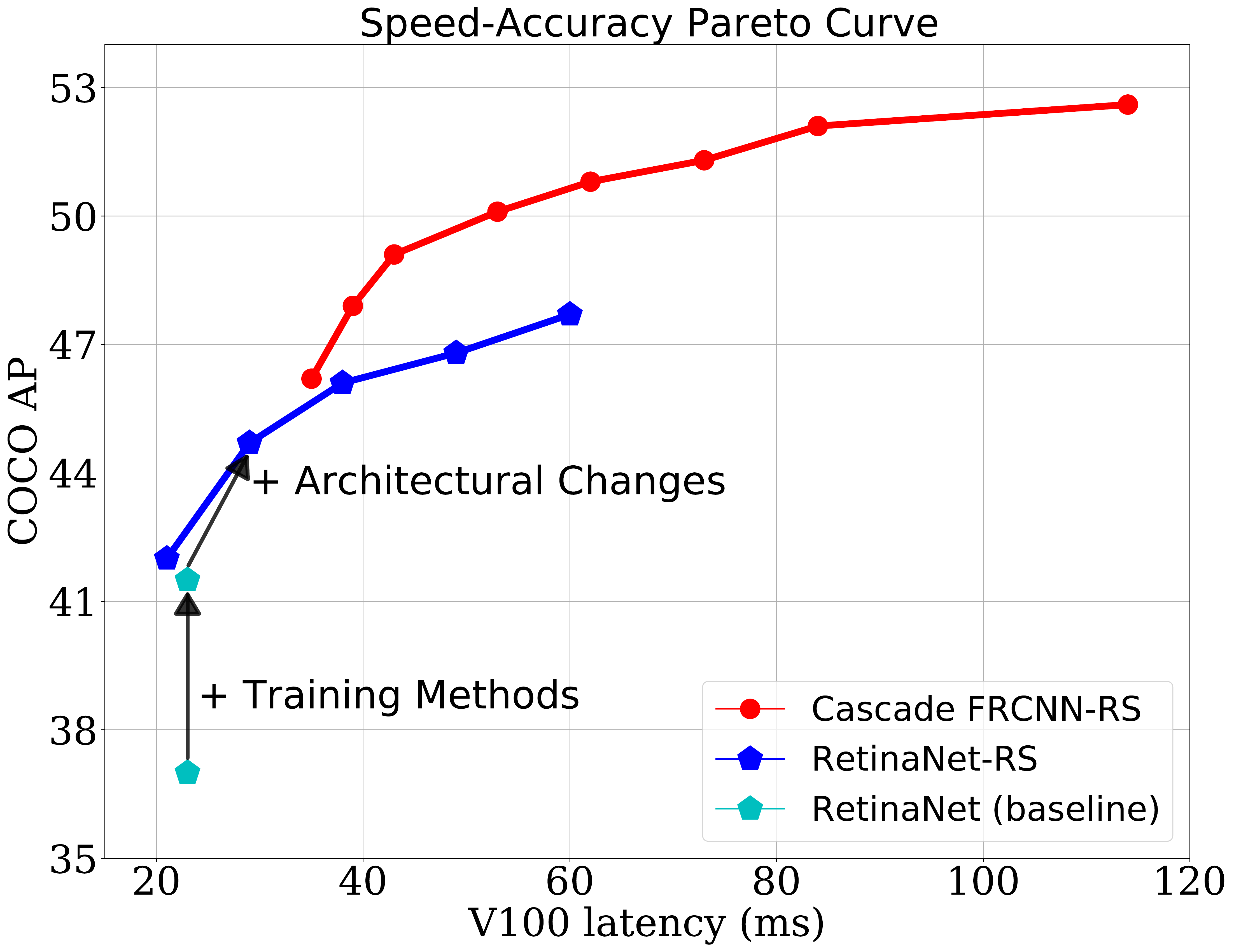}
\caption{\textbf{Performance comparisons of Vanilla RetinaNet, RetinaNet-RS and Cascade FRCNN-RS on COCO Object Detection.} Latencies are measured in \texttt{float16}. Experimental details are in Section~\ref{sec:exp_coco_box}.}
\label{fig:page1_fig}
\vspace{-0mm}
\end{figure}

\section{Introduction}
State-of-the-art object detection performance on the COCO benchmark~\cite{coco} has been pushed from 30\% AP to 58\% since the development of popular object detectors~\cite{ssd, yolo, yolov3, rcnn, fast_rcnn, fasterrcnn, mrcnn, cai2018cascade, htc, retinanet, centernet, fpn, hrnet, panet, nasfpn, spinenet, Du2020EfficientSB, efficientdet, detr, bochkovskiy2020yolov4, zhang2020resnest, copypaste,wang2021scaledyolov4, liu2021swin}. In addition to pursuing high accuracies, the research community also cares about the speed-accuracy Pareto curve of object detection systems~\cite{huang2017speed}. The performance improvements not only come from the novel model architectures proposed in the literature but also from improved scaling, and modern training and inference methods\footnote{\url{https://ai.facebook.com/blog/advancing-computer-vision-research-with-new-detectron2-mask-r-cnn-baselines/}}.
Many techniques that contribute to a large portion of performance improvements are not emphasized enough or are overlooked in the literature. Some of these details can conflate the performance of current state-of-the-art detection systems.

Our work aims to carefully study these techniques and tease apart where most of the improvements are coming from in modern detection systems. We carefully ablate the impact of the commonly used techniques in current state-of-the-art object detection and instance segmentation systems from two angles: 1) minor architectural improvements, including Squeeze-and-Excitation modules~\cite{senet}, activation functions\cite{silu} and model stem\cite{he2019bag};
2) training and inference methods, including stronger data augmentation, better model regularization\cite{dropconnect}, longer training schedules and \texttt{float16} benchmarking.

Lastly, we propose a simple yet effective model scaling method for object detection and instance segmentation. Based on our empirical results, we find that only scaling the input image resolution and backbone depth is already quite effective. We apply this strategy to RetinaNet~\cite{retinanet} and RCNN~\cite{fasterrcnn,mrcnn} models and name them RetinaNet-RS and RCNN-RS.

We evaluate our RetinaNet-RS and Cascade RCNN-RS models on the COCO dataset~\cite{coco} and the Waymo Open dataset~\cite{waymo} (WOD). Our results reveal that the above mentioned architectural changes and training/inference methods improve detection baselines by \textcolor{blue}{7.7\%} AP while reducing inference time by \textcolor{blue}{30\%}.
By further adopting the proposed scaling strategy, we present two families of detectors. In particular, our Cascade RCNN-RS model adopting a ResNet152-FPN backbone achieves 52.9\% AP on COCO at 119ms per image on a V100 GPU. Our Cascade RCNN-RS adopting a SpineNet143L backbone achieves 53.6\% AP on COCO and 71.2 AP/L1 on WOD. 


Our contributions are:

\begin{itemize}
    \item We identify the key architectural changes, training methods and inference methods that significantly improve object detection and instance segmentation systems in speed and accuracy.
    \item We highlight the key implementation details and establish new baselines for RetinaNet and Cascade RCNN models.
    \item We provide two object detection model families as strong new baselines for future research, \textbf{RetineNet-RS} and \textbf{Cascade RCNN-RS}.
    \item We explore speed-accuracy trade-off between one-stage RetinaNet and two-stage RCNN models.
\end{itemize}


\setlength{\tabcolsep}{4pt}
\begin{table}[t]
\vspace{0mm}
\begin{center}{
\begin{tabular}{l | l |l}
\toprule
 Model & AP \textcolor{gray}{(\%)} &  Latency \textcolor{gray}{(ms)} \\
 \midrule
RetinaNet baseline &  37.0 & 41\\
+ \texttt{float16} &  37.0 & 23 (\textcolor{blue}{-18})\\
+ SJ \& 350-epoch &  40.5 (\textcolor{blue}{+3.5}) & 23 \\
+ SD \& 600-epoch & 41.5 (\textcolor{blue}{+1.0}) & 23 \\
+ SiLU activation & 42.5 (\textcolor{blue}{+1.0}) & 27 (\textcolor{red}{+4}) \\
+ SE & 44.0 (\textcolor{blue}{+1.5}) & 29 (\textcolor{red}{+2})\\
+ ResNet-D & 44.7 (\textcolor{blue}{+0.7}) & 29\\
\bottomrule
\end{tabular}
}
\end{center}
\caption{\textbf{Ablation study of the modern techniques discussed in this paper.} Results are reported using a RetinaNet detector with a ResNet-50 backbone at $640\times 640$ input resolution on COCO \texttt{val2017}. SJ: scale jittering. SD: stochastic depth. SE: Squeeze-and-Excitation. ResNet-D: ResNet-D style stem. We show that there is a 7.7\% AP improvement by adopting all the techniques while reducing inference latency by 30\%.}
\label{tab:resnet_modifications}
\vspace{-0mm}
\end{table}

\section{Methodology}\label{sec:methodology}
\subsection{Modified ResNet Backbone}\label{sec:resnet_modifications}
We modify the standard ResNet~\cite{resnet} architecture in three ways to improve its performance with small computational costs. Bello~\etal~\cite{bello2021revisiting} has demonstrated Squeeze-and-Excitation module~\cite{senet} and ResNet-D stem~\cite{inceptionv2,bagoftricks} are both effective for classification model. Recent detection systems, \eg,~\cite{efficientdet, bochkovskiy2020yolov4}, show the above two changes and a non-linear activation function like Sigmoid Linear Unit activation are effective on improving detection performance.

\paragraph{\bf Squeeze-and-Excitation:} We apply the Squeeze-and-Excitation module to all  all residual blocks in the ResNet architecture. Following~\cite{senet}, one attention module is placed after the final 1$\times$1 convolutional layer, but before merging the residual with the shortcut connection. A squeeze ratio of $0.25$ is used for all experiments. 

\paragraph{\bf ResNet-D stem:} We follow~\cite{inceptionv2,bagoftricks} and modify the original ResNet stem to the ResNet-D stem. In summary, we replace the first $7\times7$ convolutional layer at feature dimension $64$ with three $3\times3$ convolutional layers at feature dimension $32, 32, 64$, respectively. The first $3\times3$ convolution has a stride of 2. Batch normalization and activation layers are applied after each convolutional layer.

\paragraph{\bf Sigmoid Linear Unit activation:} The Sigmoid Linear Unit (SiLU)~\cite{GELU}, computed as $f(x)=x\cdot \sigma(x)$, has shown promising results as a replacement of the ReLU activation. In this work, we replace all ReLU activations in the model architecture (backbone, FPN and detection heads) with the SiLU activation.

\setlength{\tabcolsep}{4pt}
\begin{table}[b]
\begin{center}{
\begin{tabular}{c | c c}
\toprule
 Model Scale & Resolution & Backbone\\
 \midrule
Scale 1 & $512\times512$ & ResNet-50 \\
Scale 3 & $640\times640$ & ResNet-50 \\
Scale 4 & $640\times640$ & ResNet-101 \\
Scale 5 & $768\times768$ & ResNet-101 \\
Scale 6 & $768\times768$ & ResNet-152 \\
\bottomrule
\end{tabular}
}
\end{center}
\caption{\textbf{RetinaNet-RS scaling method}. A simple scaling method to scale up RetinaNet detection systems by changing only the input resolution and ResNet backbone depth.}
\label{tab:scaling_ret}
\vspace{-0mm}
\end{table}

\setlength{\tabcolsep}{4pt}
\begin{table}[!ht]
\begin{center}{
\begin{tabular}{c | c c}
\toprule
 Model Scale & Resolution & Backbone\\
\midrule
Scale 1 & $512\times512$ & ResNet-50 \\
Scale 2 & $640\times640$ & ResNet-50 \\
Scale 3 & $768\times768$ & ResNet-50 \\
Scale 4 & $768\times768$ & ResNet-101 \\
Scale 5 & $896\times896$ & ResNet-101 \\
Scale 6 & $896\times896$ & ResNet-152 \\
Scale 7 & $1024\times1024$ & ResNet-152 \\
Scale 8 & $1280\times1280$ & ResNet-152 \\
Scale 9 & $1280\times1280$ & ResNet-200 \\
\bottomrule
\end{tabular}
}
\end{center}
\caption{\textbf{RCNN-RS scaling method}. A simple scaling method to scale up RCNN models by changing only the input resolution and ResNet backbone depth.}
\label{tab:scaling_cas}
\vspace{-0mm}
\end{table}

\subsection{Training and Inference Methods}\label{sec:training_benchmark_settings}

\paragraph{Strong data augmentation:} We apply horizontal flipping and image scale jittering with a random scale between [0.1, 2.0] is used as our main data augmentation strategy. The same scale jittering strategy is used and studied in~\cite{spinenet,efficientdet,copypaste}. For example, if the output image size is $640\times640$, we first resize the image to randomly between $64\times64$ and $1280\times1280$ and then pad or crop the resized image to $640\times640$.

\paragraph{Strong regularization:} We apply 4e-5 weight decay and stochastic depth~\cite{dropconnect} with 0.2 initial drop rate for model regularization. We set the drop rate for a network block based on its depth in the network. The final drop rate of one block is calculated by multiplying the initial drop rate with the block order divided by the total number of blocks.

\paragraph{Longer training schedule:} The strong data augmentation and regularization methods are combined with a longer training schedule to fully train a model to convergence. On different datasets, we keep increasing the training epochs until we find the best schedule.

\paragraph{Inference methods:} For inference we use the same square image size as training. We resize the longer side of an image to the target size and pad zeros to keep aspect ratio. We measure inference speed on a Tesla V100 GPU with batch size 1 under settings that only includes the model forward pass time and forward pass plus post-processing (\eg, non-maximum suppression) time. We report both latency measured with \texttt{float16} precision and \texttt{float32} precision. Further inference time speedup can be achieved by TensorRT optimization, which is not used in this work.

\subsection{Model Scaling Method}\label{sec:scaling_method}
We present a simple yet effective scaling method for the one-stage RetinaNet detectors and the two-satge RCNN detectors. The compounding scaling rule in EfficientDet~\cite{tan2019efficientnet,efficientdet} scales up input resolution together with model depth and feature dimensions for all model components including the backbone, FPN and detection heads. We find that only scaling up the model in input resolution and backbone depth is quite effective for most stages of the speed-accuracy Pareto curve, while being significantly simpler. We empirically control the image resolution and backbone model, then perform a grid search as shown in Table~\ref{tab:resnet_modifications} to determine the Pareto curve.

For RetinaNet, we scale up input resolution from $512$ to $768$ and the ResNet backbone depth from $50$ to $152$. As RetinaNet performs dense one-stage object detection, we find scaling up input resolution leads to large resolution feature maps hence more anchors to process. This results in a higher capacity dense prediction heads and expensive NMS. We stop at input resolution $768\times768$ for RetinaNet.
The scaling method is presented in Table~\ref{tab:scaling_ret}. We name the rescaled RetinaNet models RetinaNet-RS.

Scaling up input resolution for RCNN models is more effective than one-stage detectors. RCNN uses a two-stage object detection mechanism.
The first region proposal stage is typically lightweight and class-agnostic, thus the input resolution does not place too much overhead at the first stage.
The second stage always processes a fixed number of proposals generated from the first stage. We design the scaling method to scale up input resolution from $512$ to $1280$ and the ResNet backbone depth from $50$ to $200$. The scaling method for RCNN models is presented in Table~\ref{tab:scaling_cas} and the re-scaled model family is named as RCNN-RS.

\newcommand{\light}[1]{\textcolor{gray}{#1}}
\begin{table*}[!ht]\centering
\small
\begin{tabular}{c c  c | ccc| c | c c c}
\toprule
 \multicolumn{1}{c}{\multirow{2}{*}{Backbone}} & \multicolumn{1}{c}{\multirow{2}{*}{Detector}} & \multicolumn{1}{c|}{\multirow{2}{*}{Resolution}}& \multicolumn{3}{c|}{Latency \light{(ms)}} & \multicolumn{1}{c|}{\multirow{2}{*}{\hspace{5mm}AP\hspace{5mm}}} & \multicolumn{1}{c}{\multirow{2}{*}{AP$_{S}$}}& \multicolumn{1}{c}{\multirow{2}{*}{AP$_{M}$}} & \multicolumn{1}{c}{\multirow{2}{*}{AP$_{L}$}} \\
  &  & &
 \multicolumn{1}{c}{\text{FP16}}  & \multicolumn{1}{c}{\text{FP16$^{\dagger}$}} & \multicolumn{1}{c|}{\text{FP32}} & \multicolumn{1}{c|}{} \\
 \midrule
  ResNet50-FPN & RetinaNet & 512$\times$512&  22 & 14 & \light{28}  & 42.0 & \light{22.7} & \light{47.1}   & \light{58.4}  \\
  ResNet50-FPN & RetinaNet & 640$\times$640&  29 & 17 & \light{44}  &  44.7 &  \light{27.0}  & \light{48.5}  &  \light{60.0} \\
  ResNet101-FPN & RetinaNet & 640$\times$640&  38 & 24 & \light{57} &46.1 &  \light{28.0}  &  \light{50.3}  &  \light{62.3} \\
  ResNet101-FPN & RetinaNet & 768$\times$768&  49 & 30 & \light{76} & 46.8 &  \light{29.6} &  \light{50.6}  &  \light{62.8} \\
  ResNet152-FPN & RetinaNet & 768$\times$768&  60 & 38 & \light{87}  & 47.7 & \light{30.4}  &  \light{52.2} & \light{63.6}   \\
 \midrule
  ResNet50-FPN & Cascade FRCNN & 512$\times$512&  35 & 27 & \light{59} & 46.2 &  \light{26.2}  & \light{50.8} & \light{63.9}  \\
  ResNet50-FPN & Cascade FRCNN & 640$\times$640&  39 & 31 & \light{67} & 47.9 &  \light{30.0}  &  \light{52.3}  &  \light{64.3} \\
  ResNet50-FPN & Cascade FRCNN & 768$\times$768&  43 & 33 & \light{75}&  49.1 & \light{31.5}  &  \light{53.0}  &  \light{64.5} \\
  ResNet101-FPN & Cascade FRCNN & 768$\times$768&  53 & 41 & \light{87} & 50.1 & \light{33.4}   & \light{53.9}   & \light{65.2}  \\
  ResNet101-FPN & Cascade FRCNN & 896$\times$896&  62 & 48 & \light{100} & 50.8 & \light{34.0} & \light{54.5}   & \light{65.6}  \\
  ResNet152-FPN & Cascade FRCNN & 896$\times$896&  73 & 59 & \light{116} & 51.3 & \light{33.9}  &  \light{55.0}  & \light{65.9}  \\
  ResNet152-FPN & Cascade FRCNN & 1024$\times$1024&  84 & 70 & \light{136} & 52.1 & \light{35.3}  & \light{55.8}  & \light{66.5}  \\
  ResNet152-FPN & Cascade FRCNN & 1280$\times$1280&  114 & 96 & \light{181} & 52.6 & \light{36.7}  & \light{55.8}   & \light{66.6}  \\
 
\bottomrule
\end{tabular}
\caption{Result comparisons on COCO \texttt{val2017} of RetinaNet-RS (first group) and Cascade FRCNN-RS (second group). We report end-to-end latency including post-processing (\eg NMS) on a Tesla V100 GPU with \texttt{float16} precision (FP16) and \texttt{float32} precision (FP32). FP16$^\dagger$ represents model latency in \texttt{float16} without measuring post-processinig ops (NMS).}
\label{tab:mainresults}
\vspace{-0mm}
\end{table*}

\subsection{Detection Framework}~\label{sec:det_frameworks}
\subsubsection{RetinaNet-RS}
\label{sec:retinanet}
\paragraph{\bf Detection head:} We follow the standard RetinaNet~\cite{retinanet} head design. In brief, we use 4 $3\times3$ convolutional layers at feature dimension $256$ in the box and classification subnets before the final prediction layers. Each convolutional layer is followed by a batch norm layer and a SiLU activation. The convolutional layers are shared across all feature levels in the detection head while the batch norm layers are not shared. We place 3 anchors at aspect ratios $[1.0, 2.0, 0.5]$ at each pixel location and set the base anchor size to $3.0$. The focal loss parameters $\alpha$ and $\gamma$ are set to 0.25 and 1.5, respectively. 

\paragraph{\bf Feature extractor:} RetinaNet-RS uses the backbone, \eg, modified ResNet-50/101/152 described in Section~\ref{sec:resnet_modifications}. A standard FPN~\cite{fpn} at feature dimension $256$ is applied after the backbone to extract multi-scale features from level $P_3$ to $P_7$.


\subsubsection{Cascade RCNN-RS}\label{sec:cascade_rcnn}
We use Cascade RCNN, one of the strongest RCNN detectors, as our two-stage detection framework in this work. 
\paragraph{\bf RPN head:} For our Cascade RCNN-RS experiments we generally follow the implementation from~\cite{cai2018cascade}. For the RPN head, we use two 3$\times$3 convolutional layers at feature dimension $256$ and the same anchor settings as RetinaNet mentioned in Section~\ref{sec:retinanet}. We use $500$ proposals for training and $1000$ proposals for inference.

\paragraph{\bf Box regression head:} We use two settings for the box regression heads, one for regular-size models and one for large-size models. For regular-size models, we implement two cascaded heads with increasing IoU thresholds $0.6$ and $0.7$. Each head has 4 3$\times$3 convolutional layers at feature dimensions $256$ and one fully connected layer at feature dimension $1024$ before the final prediction layer.
We importantly note for getting good performance improvements class agnostic bounding box regression must be used. This is where for the box regression heads only 4 bounding box coordinates are predicted instead of $4 \times \text{(number of classes)}$.

\paragraph{\bf Instance segmentation head:} We use four 3$\times$3 convolutional layers and one 3$\times$3 stride-2 deconvolutional layer at feature dimension $256$ before the final prediction layer in the instance segmentation head. 

\paragraph{\bf Feature extractor:} We first study the performance of the ResNet-50/101/152/200 model family and the EfficientNet B1 to B7 model family with the regular-size Cascade RCNN framework. To scale up ResNet based models, we use the scaling method described in Table~\ref{tab:scaling_cas}. To scale up EfficientNet based models, we follow the compound scaling rule introduced in~\cite{efficientdet}. A standard FPN is attached to ResNet and EfficientNet backbones to extract $P_3$ to $P_7$ multi-scale features. To obtain the best performance, we adopt the SpineNet-143/143L backbones. The SpineNet-143L backbone uniformly scales up feature dimension of all convolutional layers in SpineNet-143 by $1.5\times$.

\begin{figure}
    \includegraphics[width=1.0\columnwidth]{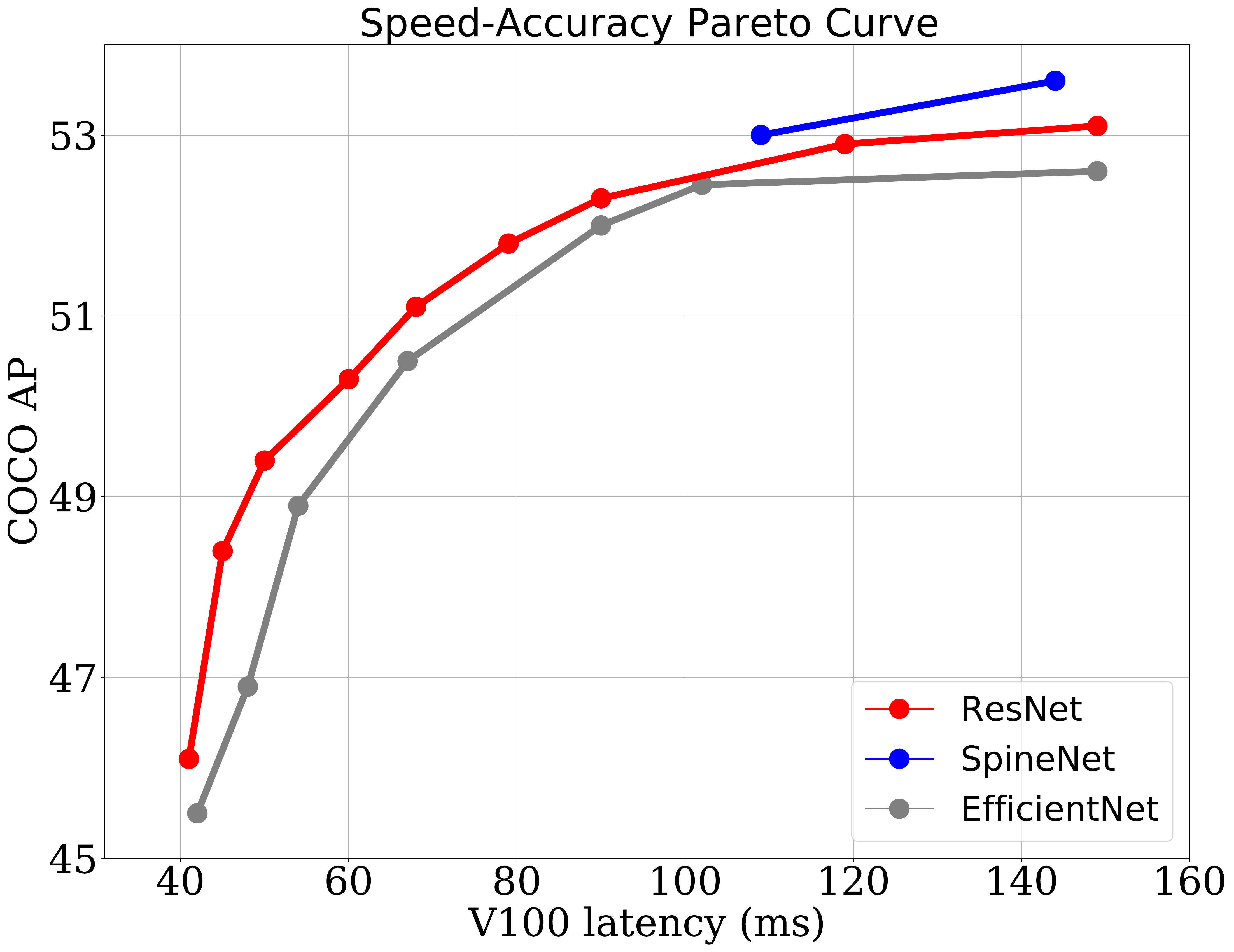}
    \caption{\textbf{Performance comparison of Cascade MRCNN-RS models adopting ResNet-FPN, EfficientNet-FPN and SpineNet backbones.} Results for all models are generated under the same settings described in Section~\ref{sec:exp_coco_inst}. We show ResNet-FPN and SpineNet backbones outperform EfficientNet-FPN backbones at all model scales.}
    \label{fig:inst_seg_curves}
\vspace*{-0mm}
\end{figure}
\begin{figure}
    \includegraphics[width=1.0\columnwidth]{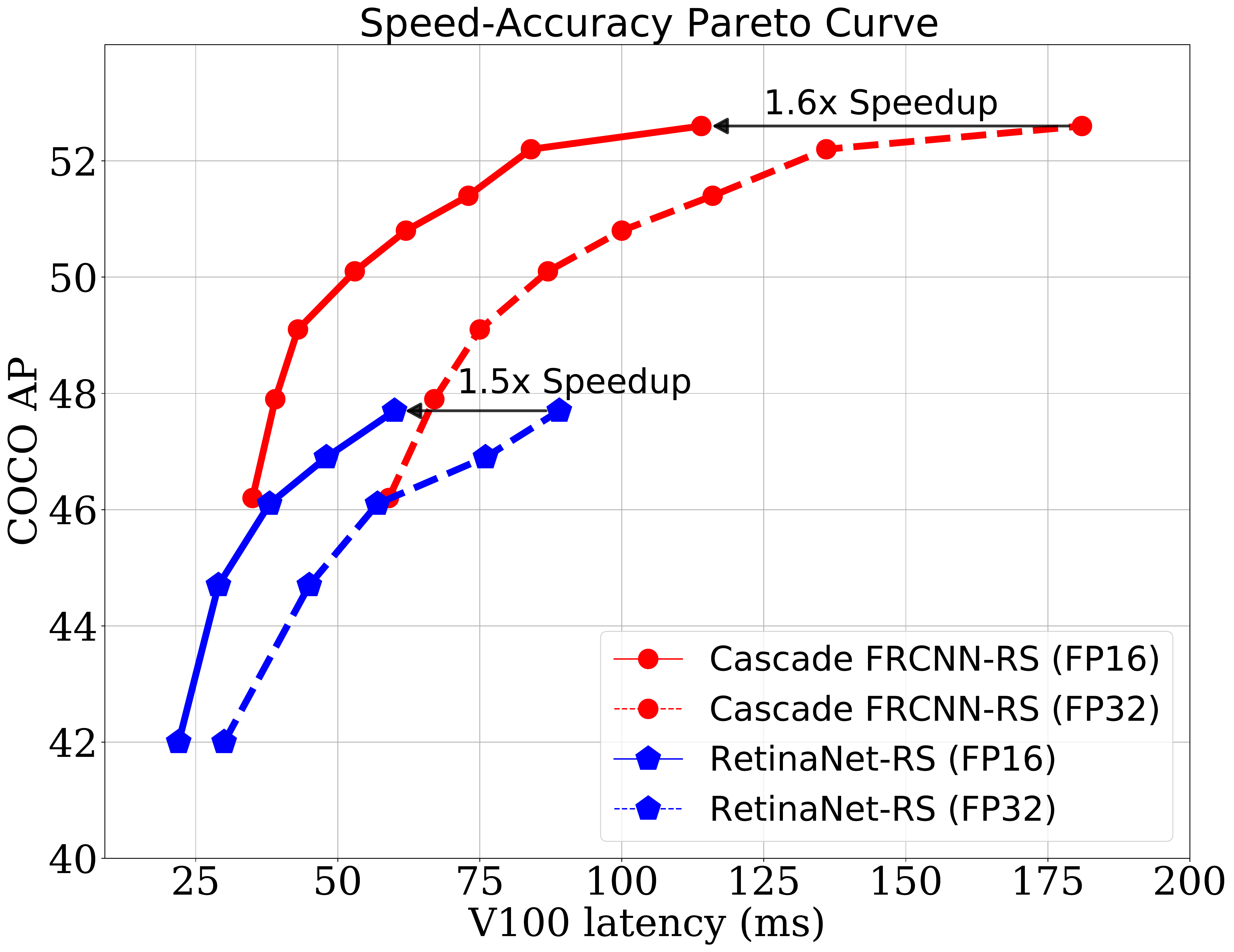}
    \caption{\textbf{Speed improvements with \texttt{float16} precision.} Inference with \texttt{float16} leads to a 1.5$\times$ to 1.7$\times$ speed boost for RetinaNet-RS and Cascade FRCNN-RS models. Latency numbers are reported on a V100 GPU.}
\label{fig:precision_comp}
\vspace{-0mm}
\end{figure}

\section{Experimental Results}\label{sec:experiments}
\subsection{Experimental settings}
\paragraph{COCO dataset:}
We first evaluate our models on the popular COCO benchmark~\cite{coco}. All models are trained on the \texttt{train2017} split and evaluated on the \texttt{val2017} split. Besides the settings described in Section~\ref{sec:training_benchmark_settings}, we generally follow~\cite{spinenet,Du2020EfficientSB,efficientdet} to train models from scratch on COCO \texttt{train2017} with synchronized batch normalization and SGD with a 0.9 momentum rate. Unless noted, all models are trained with a batch size of 256 for 600 epochs on TPUv3 devices~\cite{jouppi2017tpu}. We apply a step-wise decay learning rate schedule with an initial learning rate 0.28 that decays to $0.1\times$ and $0.01\times$ at the last 25 epoch and the last 10 epoch. A linear learning rate warm-up is applied over the first 5 epochs. Our main results for bounding box detection and instance segmentation are reported on COCO \texttt{val2017}.

\subsection{COCO Bounding Box Detection}\label{sec:exp_coco_box}
Our results of RetinaNet-RS and Cascade Faster RCNN-RS (Cascade FRCNN-RS) on the COCO bounding box detection task are presented in Table~\ref{tab:mainresults} and Figure~\ref{fig:page1_fig}.

From Figure~\ref{fig:page1_fig} we can see that the new training methods improve COCO AP by 4.5\% with no inference cost and the architectural changes improve COCO AP by another 3.2\% with insignificant computational cost. Figure~\ref{fig:page1_fig} further shows that at high-computation regime (\eg when using large model scale and large input size), the two-stage Cascade RCNN-RS models outperform the one-stage RetinaNet-RS models on the speed-accuracy pareto curve. At low-computation regime, RetinaNet-RS is more efficient than Cascade FRCNN-RS.

\begin{table*}\centering
\small
\begin{tabular}{c c  | c c| c | c c c}
\toprule
 
  \multicolumn{1}{c}{\multirow{2}{*}{Backbone}} & \multicolumn{1}{c|}{\multirow{2}{*}{Resolution}} & \multicolumn{2}{c|}{Latency \light{(ms)}} & \multicolumn{1}{c|}{\multirow{2}{*}{\hspace{5mm}AP\hspace{5mm}}} & \multicolumn{1}{c}{\multirow{2}{*}{AP$_{S}$}}& \multicolumn{1}{c}{\multirow{2}{*}{AP$_{M}$}} & \multicolumn{1}{c}{\multirow{2}{*}{AP$_{L}$}} \\
  &  &
 \multicolumn{1}{c}{\text{FP16}}  & \multicolumn{1}{c|}{\text{FP32}} & \multicolumn{1}{c|}{} \\
 \midrule
  EfficientNetB1-FPN  & 640$\times$640& 42 &\light{71}&  45.5 &  \light{24.3}  &  \light{48.8}  &  \light{65.0}  \\
  EfficientNetB2-FPN  & 768$\times$768& 48 &\light{78}&  46.9 & \light{26.4}   & \light{49.8}   &  \light{66.5}  \\
  EfficientNetB3-FPN  & 896$\times$896& 54&\light{90}& 48.9 & \light{29.2}  & \light{52.0}  & \light{67.2}   \\
  EfficientNetB4-FPN  & 1024$\times$1024& 67&\light{109}&  50.5 &  \light{30.5}  & \light{53.1}   & \light{69.2}  \\
  EfficientNetB5-FPN  & 1280$\times$1280&  90&\light{158} &  52.0 &  \light{32.3}  & \light{55.3}   & \light{69.7}   \\
  EfficientNetB6-FPN  & 1280$\times$1280&  101&\light{179}&  52.5 &  \light{33.0}  &  \light{55.4}  & \light{69.9}   \\
  EfficientNetB7-FPN  & 1536$\times$1536& 149&\light{263} & 52.6 & \light{32.2}  & \light{55.8}   &  \light{70.2} \\
 \midrule
  ResNet50-FPN  & 512$\times$512&  41&\light{69}&  46.1 &  \light{23.7}  &  \light{50.1}  & \light{67.1}   \\
  ResNet50-FPN  & 640$\times$640&  45&\light{77}&  48.4 &  \light{27.4}  & \light{51.5}   & \light{67.9}    \\
  ResNet50-FPN  & 768$\times$768&  50&\light{85}&  49.4 &  \light{29.4}  & \light{52.6}   & \light{68.3}  \\
  ResNet101-FPN & 768$\times$768&  60&\light{97}&  50.3 & \light{29.7} & \light{53.9} &  \light{69.7}   \\
  ResNet101-FPN & 896$\times$896&  68&\light{109}&  51.1 &  \light{31.4}  & \light{54.3}  & \light{69.8}    \\
  ResNet152-FPN & 896$\times$896&  79&\light{125}&  51.8 &  \light{32.0}  & \light{55.1} & \light{70.0} \\
  ResNet152-FPN & 1024$\times$1024&  90&\light{148}&  52.4 &  \light{32.9}  &  \light{55.3}  & \light{70.0}  \\
  ResNet152-FPN & 1280$\times$1280&  119&\light{191}&  52.9 &  \light{33.5}  &  \light{56.7}  &  \light{70.3}  \\
  ResNet200-FPN &  1280$\times$1280&  149&\light{232}&  53.1 & \light{33.9}  & \light{56.2}   & \light{70.3}  \\
  \midrule
  SpineNet143 & 1280$\times$1280&  109&\light{175}&  53.0 &  \light{33.8}  & \light{55.8} & \light{70.5}  \\
  SpineNet143L & 1280$\times$1280&  144&\light{234}&  53.6 & \light{34.5} & \light{56.7} & \light{70.6}  \\
  
\bottomrule
\end{tabular}
\vspace*{-0mm}
\caption{\textbf{Result comparisons on COCO \texttt{val2017} among Cascade MRCNN-RS models adopting ResNet-FPN, SpineNet and EfficientNet-FPN backbones.} All results are generated in the same codebase. We report end-to-end latency including NMS on a Tesla V100 GPU with \texttt{float16} precision (FP16) and \texttt{float32} precision (FP32).}
\label{tab:instace_seg_reuslts}
\vspace{-0mm}
\end{table*}

\subsection{COCO Instance Segmentation}\label{sec:exp_coco_inst}
Cascade Mask RCNN-RS (Cascade MRCNN-RS) is used as our two-stage instance segmentation framework. We compare three backbones: ResNet-FPN, EfficeintNet-FPN and SpineNet. All models are trained and evaluated in the same codebase and benchmarked on a Tesla V100 GPU under the same settings. Results are presented in Table~\ref{tab:instace_seg_reuslts} and Figure~\ref{fig:inst_seg_curves}.

Adopting the same Cascade MRCNN-RS framework and trained under the same settings, ResNet-FPN backbones are able to achieve a better speed-accuracy Pareto curve than the EfficientNet-FPN backbones at all model scales. The Cascade MRCNN-RS model adopts a ResNet200-FPN backbone at input resolution 1280 achieves 53.1\% AP. By replacing the backbone with SpineNet-143L, we further improve the AP to 53.6\% with a slightly faster speed.

\section{Understanding Performance Improvements}
\subsection{Effectiveness of Modern Techniques}
We conduct ablation studies for the modern training methods and architectural modifications with a RetinaNet-RS model that adopts a ResNet50-FPN backbone at 640 input resolution. As shown in Table~\ref{tab:resnet_modifications}, training with large scale jittering for 350 epochs improves AP by 3.5\%. Stronger model regularization and a longer 600-epoch training schedule improves AP by 1.0\%. The modern training methods improve AP by 4.5\% without introducing any computational costs. For architectural modifications, replacing ReLU with SiLU activation improves AP by 1.0\%, Plugging in Squeeze-and-Excitation modules improve AP by 1.5\% and adopting the ResNet-D stem improves AP by another 0.7\%. The three modifications improve AP by 3.2\% while introducing insignificant computational costs. The results are also presented in Figure~\ref{fig:page1_fig}.

\setlength{\tabcolsep}{4pt}
\begin{table*}[!h]
\begin{center}{
\begin{tabular}{c | c c  c c c c}
\toprule
Backbone \textbackslash Resolution & 512 & 640 & 768 & 896 & 1024 & 1280 \\
\midrule
R50-FPN & 46.1 \light{(69)} & 48.4 \light{(77)} & 49.4 \light{(85)} & 50.0 \light{(94)} & 50.2 \light{(106)} & 50.8 \light{(132)} \\
R101-FPN & 47.5 \light{(75)} & 49.3 \light{(86)} & 50.3 \light{(97)} & 51.1 \light{(109)} & 51.6 \light{(123)} & 51.9 \light{(160)} \\
R152-FPN & 47.8 \light{(83)} & 49.7 \light{(96)} & 50.6 \light{(109)} & 51.8 \light{(125)} & 52.3 \light{(148)} & 52.9 \light{(193)} \\
R200-FPN & 48.3 \light{(92)} & 50.4 \light{(109)} & 51.5 \light{(127)} & 52.2 \light{(148)} & 52.6 \light{(174)} & 53.1 \light{(232)} \\
\bottomrule
\end{tabular}
}
\end{center}
\caption{\textbf{ResNet depth vs input resolution.} We report COCO \texttt{val2017} AP of different ResNet backbones and the corresponding inference latency on V100 GPU (numbers in parentheses) with \texttt{float32} precision. All models adopt the Cascade MRCNN-RS detector.}
\label{tab:resnet_resolution}
\vspace{-0mm}
\end{table*}


\begin{figure}[!h]
    \includegraphics[width=1.0\columnwidth]{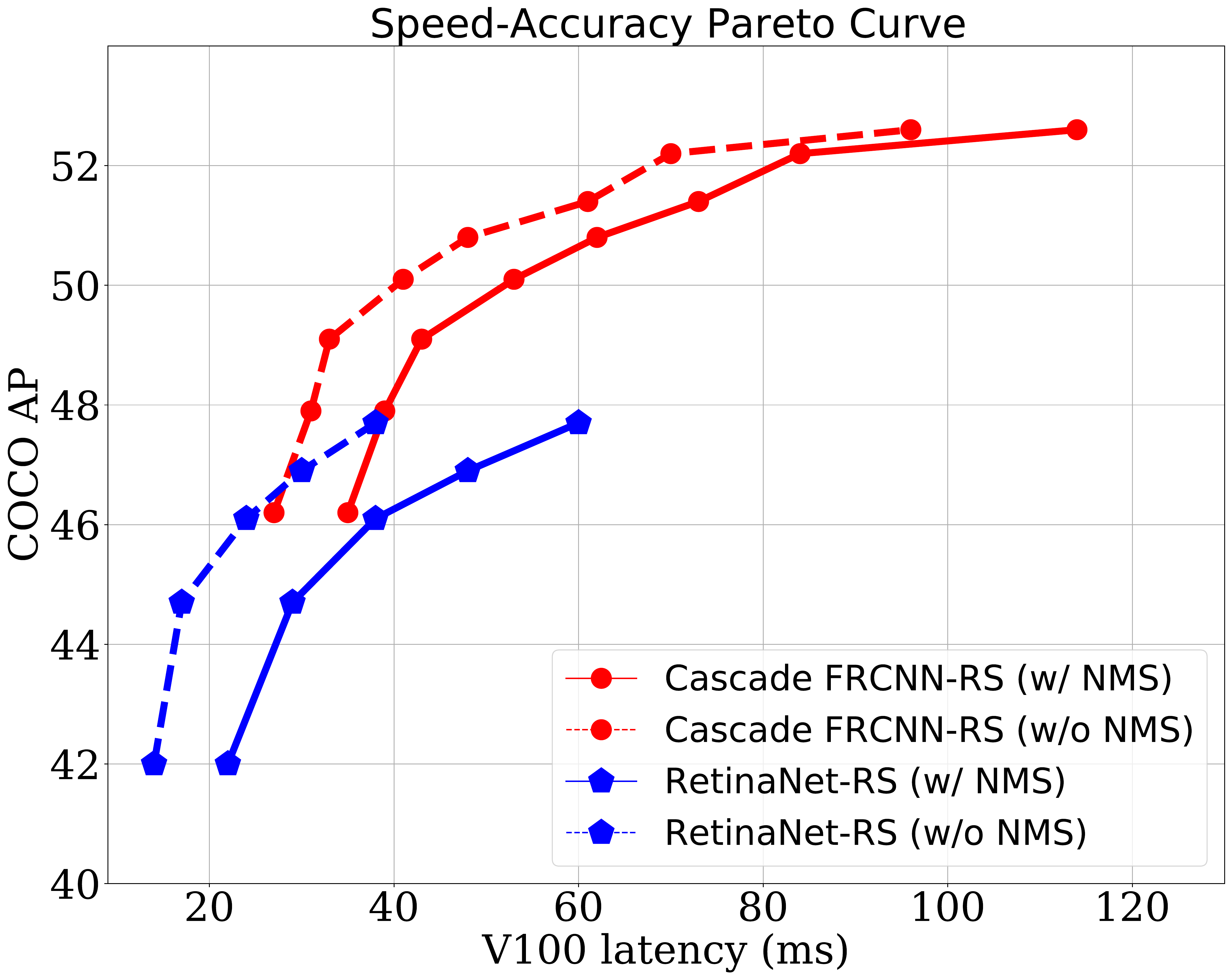}
    \caption{\textbf{Speed comparisons with or without measuring post-processing ops for RetinaNet-RS and Cascade FRCNN-RS.} Latency numbers are reported on a V100 GPU.}
\label{fig:nms_comp}
\vspace{-0mm}
\end{figure}

\subsection{Latency benchmarking settings}
\paragraph{Precision float16:} Inference in 16-bit floating-point types make model run faster and use less memory. This subsection compares RetinaNet-RS and Cascade FRCNN-RS models using \texttt{float32} precision and \texttt{float16} precision for model inference. By casting model weights and input images from \texttt{float32} to \texttt{float16} precision, the speed of model forward pass can be boosted to 1.5$\times$ to 1.7$\times$. Inference latency and speed improvements of all models are presented in Figure~\ref{fig:precision_comp} and Table~\ref{tab:mainresults},~\ref{tab:instace_seg_reuslts}.
\paragraph{Post-processing:} Post-processing ops such as NMS takes a non-negligible portion of latency. We show the speed of our RetinaNet-RS and Cascade FRCNN-RS models with and without measuring post-processing ops in Table~\ref{tab:mainresults} and the pareto curve comparisons in Fig.~\ref{fig:nms_comp}.

\subsection{ResNet Depth vs Input Resolution}
This section analyzes the efficiency of the proposed scaling method that scales model depth and input resolution at the same time. We present the performance of Cascade MRCNN-RS models that adopt ResNet-50/101/152/200 backbones and trained at input resolution 512, 640, 768, 896, 1024 and 1280. We show that a better scaling efficiency can be achieved when scaling model depth and input resolution together. At low input resolutions (\ie 512 or 640), the ResNet-50 backbone is more effective than deeper ResNet variants. At slight larger input resolutions (\ie, 768 or 896), the ResNet-101 backbone is more effective. At high input resolutions (\ie, 1024 or 1280) the ResNet-152 backbone is the most effective choice. Further scaling model depth to ResNet-200 does not improve the speed-accuracy Pareto curve. The results are shown in Table~\ref{tab:resnet_resolution}.

\setlength{\tabcolsep}{4pt}
\begin{table}[!h]
\begin{center}{
\begin{tabular}{c | c c c c}
\toprule
 Jitter Scale & 90-ep & 200-ep & 400-ep & 600-ep\\
 \midrule
\text{[1.0, 1.0]} & 41.9 & 39.3 & 38.4 & 35.7 \\
\text{[0.8, 1.2]} & 43.8 & 45.0 & 44.6 & 44.3\\
\text{[0.1, 2.0]} & 43.9 & 47.0 & 47.6 & 47.9\\
\bottomrule
\end{tabular}
}
\end{center}
\caption{\textbf{Jitter scales vs training epochs.} We show that model performance significantly benefits from using aggressive scale jittering with longer training schedules.}
\label{tab:jitter_epoch}
\vspace{-0mm}
\end{table}

\setlength{\tabcolsep}{4pt}
\begin{table*}[h]
\begin{center}{
\begin{tabular}{c  | c c | c c}
\toprule
 Model & Train Resolution & Eval Resolution & AP\slash L1 & AP\slash L2\\
 \midrule
 RetinaNet-RS & 896$\times$1664 & 896$\times$1664 & 64.5 & 56.1 \\
\midrule
 FRCNN-RS & 896$\times$1664 & 896$\times$1664 & 68.1 & 60.0\\
 FRCNN-RS & 896$\times$1664 & 1024$\times$1920 & 69.5 & 61.2\\
 FRCNN-RS 
 & 896$\times$1664 & 1280$\times$2400 & 70.0 & 63.4\\
\midrule
 Cascade FRCNN-RS & 896$\times$1664 & 896$\times$1664 & 68.6 & 60.7\\
 Cascade FRCNN-RS & 896$\times$1664 & 1024$\times$1920 & 70.5 & 61.8\\
 Cascade FRCNN-RS & 896$\times$1664 & 1280$\times$2400 & 71.2 & 62.9\\

\bottomrule
\end{tabular}
}
\end{center}
\caption{\textbf{Results on Waymo Open Dataset.} We evaluate different detectors adopting the SpineNet143L backbone.}
\label{tab:waymo}
\end{table*}


\subsection{Scale Jittering vs Training Schedule}
Applying large scale jittering augmentation potentially results in more unique training samples during the training procedure, which allows the model to benefit from a longer training schedule. We empirically show that the best model performance can be achieved by enabling large scale jittering to train models for an up to 600 epoch schedule. To study the effectiveness, we compare three scale jittering setups, [0.1, 2.0], [0.8, 1.2], and [1.0. 1.0], on 90-epoch, 200-epoch, 400-epoch, and 600-epoch model training schedules. The results are shown in Table~\ref{tab:jitter_epoch}.

\subsection{Impact of Non-maximum Suppression}
In this paper, the detection frameworks (RetinaNet and R-CNN) employ non-maximum suppression (NMS) to remove duplicate detection. It becomes a bottleneck as we keep improving train the training technologies and model architecture. Figure~\ref{fig:page1_fig} and~\ref{fig:nms_comp} compares the inference time with and without the NMS operation. The non-maximum suppression takes about ~40\% of inference time in RetinaNet and 15-30\% of inference time in Cascade Faster R-CNN. A well optimized software library, \eg, TensorRT, or the detectors without NMS, \eg,~\cite{carion2020endtoend,centernet}, can save the computation overhead.


\section{Conclusion}\label{sec:conclusion}
In this work, we dissects the performance improvements coming from minor architecture modifications and training/inference methods. We design a family of models using a simple scaling strategy that only changes the backbone capacity and image resolution. We find the speed-accuracy Pareto curve is already a strong baseline to recent object detection systems. We hope this study will help the research community understand the bottleneck and performance improvements of modern object detection systems.

\appendix 

\section{Waymo Open Dataset}
\paragraph{Experment setup:}
We conduct experiments on the Waymo Open Dataset~\cite{waymo_open_dataset}, which is a large-scale dataset for autonomous driving. The dataset comprises 798~training sequences and 202~validation sequences. Each sequence spans 20~seconds and is densely labeled at 10~frames per second with camera object detection and tracks. We evaluate our models on the 2D Detection Task. 

We train all models on the \texttt{train} set and report the metrics on the \texttt{test} set. The models are pre-trained on the JFT and COCO datasets and finetuned on the Waymo Open Dataset for 6 epochs with using a cosine decay learning rate with an initial learning rate of 0.08. All other hyper-parameters are the same as the models trained on the COCO dataset.

\paragraph{Results:}
We evaluate our improved baselines with SpineNet143L as the backbone on the Waymo Open Dataset (WOD) 2D detection task~\cite{waymo_open_dataset}. Compared to the COCO dataset, WOD includes more small objects with heavy occlusions in the urban driving scene. We show the results in Table~\ref{tab:waymo}. FRCNN performs better than RetinaNet with 3.6\% AP/L1 and 3.9\% AP/L2 difference. The reason might be that there are many frames with little or no instance, and single stage detectors could suffer more from the data imbalance even with the help of the focal loss. To obtain the best performance, we adopt Cascade FRCNN and apply $7$ convolutional layers instead of $4$ in each head and attach one more cascaded head with an IoU threshold $0.8$. The performance is improved with 0.5\%-1.2\% AP on different resolutions compared to the 2-stage FRCNN. This demonstrates that the improved baselines could also generalize to data of different domains (e.g. self-driving cars).


{\small
\bibliographystyle{ieee_fullname}
\bibliography{arxiv}
}

\end{document}